\documentclass{article}
\usepackage{iclr2027_conference,times}
\usepackage{amsmath,amsfonts,bm}

\def\eqref#1{equation~\ref{#1}}

\def\1{\bm{1}}

\DeclareMathAlphabet{\mathsfit}{\encodingdefault}{\sfdefault}{m}{sl}
\SetMathAlphabet{\mathsfit}{bold}{\encodingdefault}{\sfdefault}{bx}{n}

\renewcommand{\eqref}[1]{(\ref{#1})}
\usepackage{graphicx}
\usepackage{amsmath,amssymb}
\usepackage{mathtools}
\usepackage{booktabs}
\usepackage{xcolor}

\usepackage{xspace}

\usepackage{placeins}

\usepackage{hyperref}
\usepackage{url}

\title{Every View Counts:\\ View-Consistent Panoptic Quality\\ for Multi-view Panoptic Segmentation}

\author{Youngmin Lee\thanks{Equal contribution.}\hspace{0.35em}, Byungha Ko\footnotemark[1]\hspace{0.35em}, Guhnoo Yun \& Dong Hwan Kim\thanks{Corresponding author.} \\
Korea University \\
Korea Institute of Science and Technology \\
\texttt{\{aidi0724, rhqo123, doranlyong, gregorykim\}@kist.re.kr}
}

\iclrfinalcopy
\begin{document}

\maketitle

\begin{abstract}
Multi-view panoptic segmentation assigns a semantic class and a scene-level instance ID to every pixel of an unordered set of images, and recent feed-forward 3D models predict these labels for the input views in a single forward pass.
Their predictions, however, have been evaluated with the scene-level PQ (PQ$^{\text{scene}}$) borrowed from per-scene optimization methods, typically on rendered held-out views.
PQ$^{\text{scene}}$ tiles all views of a scene into a single image, so that a missed appearance or a change of ID lowers the score of the matched pair only in proportion to its area.
We propose View-Consistent Panoptic Quality (VC-PQ), which extends PQ from a single image to a set of input views, counts equally every view in which an instance is visible, and penalizes a prediction that is not visible in the same views as its ground truth.
A decomposition of VC-PQ attributes the score a method loses to mask accuracy, view consistency, and the matching threshold.
A single additional parameter recovers the area weighting of tiling for comparison.
Under a fixed evaluation protocol on ScanNet++ and ScanNetv2, recent feed-forward methods are evaluated with VC-PQ and PQ$^{\text{scene}}$, and the decomposition shows where each of them loses its score.
Controlled perturbations of the ground truth show that VC-PQ responds to the number of views in which an instance is missed or changes ID, whereas PQ$^{\text{scene}}$ responds to their area.
The aim of this work is to make view consistency part of the evaluation of multi-view panoptic segmentation, with VC-PQ reported alongside PQ$^{\text{scene}}$.
The evaluation code and the input sets of the protocol are available at \url{https://github.com/Renem24/vcpq}.
\end{abstract}

\begin{figure}[t]
\centering
\includegraphics[width=\textwidth]{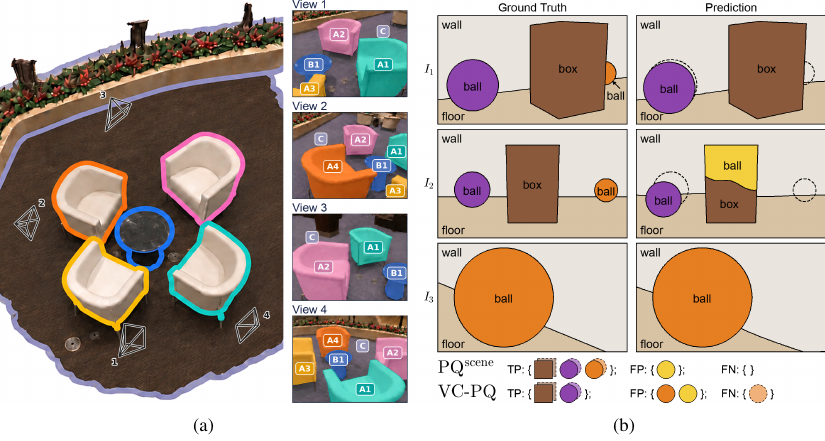}
\caption{(a) Multi-view panoptic segmentation: every view of a scene is labeled with a class and a scene-level instance ID. (b) A toy example of an input set of three views: the orange ball is predicted only in the view in which it appears large, so PQ$^{\text{scene}}$ matches it by area whereas VC-PQ counts the two missed views and leaves it unmatched.}
\label{fig:teaser}
\end{figure}
\section{Introduction}
\label{sec:intro}

Multi-view panoptic segmentation assigns a semantic class and a scene-level instance ID to every pixel of the images captured from several views of a scene, so that one object instance carries the same ID in every view in which it appears (Fig.~\ref{fig:teaser}a).
Unlike in video segmentation and object tracking, the input views are captured in an arbitrary order, with no continuity from one view to the next, and therefore this consistency of identity across views is the demanding part of the task.
The chairs in a lounge, for example, often look alike (Fig.~\ref{fig:teaser}a), so when the viewpoint changes, appearance alone may not tell whether the chairs in different views are the same instance.

Per-scene optimization methods lift the outputs of a 2D segmentation model into a 3D scene representation that is optimized for each scene given precomputed camera poses, and the shared representation enforces label consistency across views~\citep{siddiqui2023panoptic,contrastivelift,gaussiangrouping}.
Recent feed-forward 3D reconstruction models~\citep{dust3r,wang2025vggt} predict the 3D geometry of a scene from the images alone, without precomputed camera poses.
Building on their geometry-aware features, feed-forward methods for multi-view segmentation predict instance or panoptic labels with consistent IDs for the input views~\citep{panst3r,eps3d,segvggt,iggt}.

How these predictions are evaluated, however, has not followed the task.
Predictions for this task have commonly been evaluated with the scene-level Panoptic Quality (PQ$^{\text{scene}}$) of Panoptic Lifting~\citep{siddiqui2023panoptic}, which applies PQ~\citep{kirillov2019panoptic} to held-out views rendered from the per-scene representation, and feed-forward methods have adopted this protocol~\citep{panst3r,eps3d} although they predict on the input views.
PQ$^{\text{scene}}$ moreover tiles all views of a scene into a single image before applying PQ, so that a missed appearance, or a change of ID, in one view lowers the score of the matched pair only in proportion to the area of that appearance (Fig.~\ref{fig:teaser}b).
The same instance appears at a smaller scale in a view captured farther from it and covers fewer pixels in a view where other objects occlude it, so the weight a view carries in its score depends largely on the camera pose within the scene.

We therefore propose View-Consistent Panoptic Quality (VC-PQ), which extends PQ from a single image to a set of input views and scores each instance jointly by the accuracy of its masks and by whether it is visible in the same views as its ground truth, with every such view weighted equally.
Apart from a visibility threshold that disregards noisy appearances of only a few pixels, VC-PQ reduces to PQ for a single view.
A decomposition of the score lost under VC-PQ attributes it to mask accuracy, view consistency, and the matching threshold, and a one-parameter family of pair scores isolates the area weighting of tiling.

We evaluate both metrics under a fixed protocol on ScanNet++ and ScanNetv2 that draws input sets of several sizes from every scene.
Each method is evaluated with its native post-processing, and with an adapter where it does not output panoptic labels.
Controlled perturbations of the ground truth show that VC-PQ responds to the number of views in which an instance is missed or changes ID, whereas PQ$^{\text{scene}}$ responds to their area.
As the input set grows, PQ$^{\text{scene}}$ rises whereas VC-PQ falls on recent feed-forward methods, and the one-parameter family traces the fall to the equal weighting of views.

In summary, our contributions are as follows.
\begin{itemize}
\item A formulation of multi-view panoptic segmentation on the input views, with two requirements that a metric for this task should meet (Sec.~\ref{sec:problem}).
\item View-Consistent Panoptic Quality, which weights every view in which an instance is visible equally, penalizes a prediction that is not visible in the same views as its ground truth, and reduces to PQ for a single view, with a one-parameter family that isolates the area weighting of tiling (Sec.~\ref{sec:vcpq}).
\item An evaluation protocol on ScanNet++ and ScanNetv2, under which VC-PQ and PQ$^{\text{scene}}$ are compared on recent feed-forward methods and their difference is attributed to each property of PQ$^{\text{scene}}$ (Sec.~\ref{sec:eval_protocol}--\ref{sec:gamma_sweep}).
\item A decomposition of the lost score by mask accuracy, view consistency, and the matching threshold, which separates the errors within single images from those that arise only across views (Sec.~\ref{sec:vcpq_decomp}, Sec.~\ref{sec:error_decomp}).
\end{itemize}

\section{Related Work}
\label{sec:related}

\subsection{Multi-view Panoptic Segmentation}
\label{sec:rw_multiview}

Methods that assign semantic, instance, or panoptic labels to the views of a scene follow one of two approaches, per-scene optimization or feed-forward prediction.
Per-scene methods lift the predictions of an external 2D segmenter into a 3D scene representation, a neural field or 3D Gaussians, that is optimized for each scene given precomputed camera poses, and evaluate the labels rendered from it~\citep{siddiqui2023panoptic,contrastivelift,zhu2024pcflift,gaussiangrouping,liu2025mvcpsu,dhiman2026uniclift}.

Recent feed-forward 3D reconstruction models have progressively removed both requirements.
DUSt3R and its extensions~\citep{dust3r,leroy2024mast3r,cabon2025must3r} predict point maps without camera poses, and VGGT~\citep{wang2025vggt} predicts camera poses, depth maps, and point maps for a whole set of views in one forward pass.
Feed-forward segmentation methods are built on these reconstruction models as a geometry backbone, predict the labels for the input views without a per-scene optimization, and differ in how an instance identity is derived from the geometry-aware features of the backbone.
One group predicts a per-pixel or per-Gaussian instance embedding and clusters it into instances at inference, so that an ID is a cluster label and holds across views only as far as the clustering keeps an instance in one cluster~\citep{iggt,eps3d,unite,jiang2026instancesplat}.
The other shares a set of object queries across all views, so that an ID is a query index and is the same in every view by construction, and which query covers an instance in each view is left to a post-processing selection of masks or to training alone~\citep{panst3r,xu2025siu3r,barberteguy2026pano3d,segvggt,fast3dis}.
Among the feed-forward methods, PanSt3R, EPS3D, Pano3D, and SIU3R produce panoptic labels, and PanSt3R and EPS3D report panoptic quality with the scene-level protocol of per-scene methods, at novel views rendered from 3D Gaussians rather than on the input views~\citep{panst3r,eps3d}.

\subsection{Evaluation Metrics}
\label{sec:rw_metrics}

Panoptic Quality (PQ)~\citep{kirillov2019panoptic} scores a single image by matching predicted and ground-truth segments one-to-one and counting every segment equally regardless of its area.
Panoptic Lifting~\citep{siddiqui2023panoptic} extends it to a scene as PQ$^{\text{scene}}$ by tiling the views of a scene into one image and reports it at held-out test views, and later per-scene and feed-forward methods adopt this metric~\citep{contrastivelift,zhu2024pcflift,liu2025mvcpsu,dhiman2026uniclift,panst3r}.
In video panoptic segmentation, PQ is extended by matching tubes over consecutive frames, as in VPQ~\citep{kim2020vps,qiao2021vipdeeplab}, or by combining an association term with a segmentation term without matching, as in STQ~\citep{weber2021step,mei2022waymo,aygun20214d}; both pool the pixels of several frames, over a window for VPQ and over the whole sequence for STQ.

For the consistency of an instance across views, tracking evaluation offers a form that PQ does not have.
It scores a given target by its overlap in each frame, averaged with equal weight over the frames, and by the fraction of frames in which it is tracked~\citep{cehovin2016measures,huang2021got10k}.
Recent multi-view methods score a given instance in the same form, by its mean per-view IoU and by whether it is tracked in every view~\citep{iggt,zhu2026g2tam}.
Others match instances by IoU without classes and report a mean IoU and an F-score~\citep{eps3d,zhu2025unifiedlift}, and with the instance given or the class dropped, none of these scores panoptic labels with a class-wise matching and a count of unmatched instances.
VC-PQ (Sec.~\ref{sec:vcpq}) keeps the matching and unmatched terms of PQ and counts every view in which a matched pair is visible equally, in the form of these measures.

\section{Problem Setting}
\label{sec:problem}

\subsection{Task formulation}
\label{sec:problem_formulation}

Given $N$ RGB images $I_1, \dots, I_N$ of a static scene, captured from $N$ views in an arbitrary order, the task is to predict for each view a panoptic label map $Y_i$.

The panoptic label map $Y_i$ assigns every pixel $u$ of $I_i$ a panoptic label $Y_i(u) = (c, z)$~\citep{kirillov2019panoptic}, consisting of a semantic class $c$ and an instance ID $z$.
Instance IDs are defined at the scene level.
That is, pixels sharing an ID, whether in the same or in different views, belong to one object instance and carry the same class.
We refer to an instance by its ID and write $c(\cdot)$ for its class and $M_i(\cdot)$ for its mask in image $I_i$, so that $M_i(z) = \{\, u \mid Y_i(u) = (c(z), z) \,\}$, which is empty when the instance does not appear in $I_i$.
Each stuff class is treated as a single instance.

Ground-truth instances are denoted by $g$ and predicted instances by $p$, with their masks $M_i(g)$ and $M_i(p)$ defined from the respective label maps in the same way.
An ideal prediction places predicted and ground-truth instances in one-to-one correspondence such that every corresponding pair $(p, g)$ satisfies $c(p) = c(g)$ and $M_i(p) = M_i(g)$ for every image $I_i$.

\subsection{Evaluation requirements and existing metrics}
\label{sec:pqscene_mismatch}

Evaluating multi-view panoptic segmentation means measuring how closely predicted instances approach the ideal correspondence of Sec.~\ref{sec:problem_formulation} across the input views.
A metric that captures this ideal should therefore (i) reflect the mask accuracy of an instance in every view in which it appears, independent of its apparent size in that view, and (ii) penalize a predicted instance by how far the set of views in which it appears departs from that in which its ground-truth counterpart appears.

For a single image $I_i$, Panoptic Quality (PQ)~\citep{kirillov2019panoptic} scores a predicted instance $p$ against a ground-truth instance $g$ of the same class by the intersection over union (IoU) of their masks, $\mathrm{IoU}_i(p,g) = \lvert M_i(p) \cap M_i(g) \rvert / \lvert M_i(p) \cup M_i(g) \rvert$, and matches predicted and ground-truth instances one-to-one among the pairs with $\mathrm{IoU}_i(p,g) > 0.5$, which makes the matching unique.
Matched pairs, unmatched predicted instances, and unmatched ground-truth instances are counted over the dataset as true positives (TP), false positives (FP), and false negatives (FN).
PQ then divides the summed IoU of the matched pairs, each taken in the image $i$ that contains the pair, by the number of matches with a penalty of one half for every unmatched instance,
\begin{equation}
  \mathrm{PQ} = \frac{\sum_{(p,g) \in \mathrm{TP}} \mathrm{IoU}_i(p,g)}{\lvert \mathrm{TP} \rvert + \tfrac{1}{2}\lvert \mathrm{FP} \rvert + \tfrac{1}{2}\lvert \mathrm{FN} \rvert},
  \label{eq:pq}
\end{equation}
computed for each class independently and averaged over classes.

Panoptic Lifting~\citep{siddiqui2023panoptic} introduced the scene-level PQ (PQ$^{\text{scene}}$), which recent feed-forward methods also report~\citep{panst3r,eps3d}.
It tiles the predictions and ground truth of all test views of a scene into a single image and applies standard PQ, so that a pair is scored by the IoU of the concatenated masks,
\begin{equation}
  \mathrm{IoU}_{\text{concat}}(p,g) = \frac{\sum_i \lvert M_i(p) \cap M_i(g) \rvert}{\sum_i \lvert M_i(p) \cup M_i(g) \rvert},
  \label{eq:iou_concat}
\end{equation}
where the sums run over the test views of the scene.
The official implementation further discards, before matching, every class that covers less than 0.5\% of the labeled pixels of the scene in either the prediction or the ground truth.
The feed-forward methods that report PQ$^{\text{scene}}$ evaluate under this novel-view protocol, by feeding rendered held-out views to the network or by rendering their labels from a 3D field~\citep{panst3r,eps3d}.
We evaluate their predictions on the input views, as the task requires.

An input set, like a video clip, contains several views of one scene in which an instance keeps its ID from view to view, which suggests Video Panoptic Quality (VPQ) as a metric for this setting.
VPQ~\citep{kim2020vps,qiao2021vipdeeplab} matches predicted and ground-truth tubes within a window of $k$ consecutive frames and averages the score over several window lengths.
Within a window, a pair of tubes is scored by the same concatenated IoU of Eq.~\eqref{eq:iou_concat}, with the sums running over the frames of the window.

In this setting, three properties of PQ$^{\text{scene}}$ no longer match the task (Fig.~\ref{fig:pqscene_props}, Appendix~\ref{app:comparison}).
First, Eq.~\eqref{eq:iou_concat} sums pixel areas over views, so that each view weighs in proportion to the area the pair covers in it.
Accurate segmentation where an instance appears large then compensates for a complete miss where it appears small, as for the orange ball of Fig.~\ref{fig:teaser}b, and requirement (i) is not met.
A view in which only one of the two appears has an empty intersection and adds its area to the denominator alone, so a missed or spurious appearance lowers Eq.~\eqref{eq:iou_concat} only in proportion to the area it covers in that view, and requirement (ii) is not met either.
Both requirements thus turn on the same area weighting, which PQ$^{\text{scene}}$ and VPQ share.

Second, PQ$^{\text{scene}}$ yields one value per scene, and results for a dataset are reported as the average of these values across scenes~\citep{panst3r,eps3d}.
The scene thus plays for PQ$^{\text{scene}}$ the role that the image plays for PQ, yet PQ does not average over its unit, since the counts of TP, FP, and FN are pooled over all images before PQ is computed for each class~\citep{panopticapi}.
Averaging scene values instead gives every scene the same weight, so that a scene with only a few instances, all of them found, enters the average with a value close to the maximum.
For this task, the labels of the $N$ input views are predicted together, and the input set, rather than the scene, therefore takes the place of the single image of PQ.

Third, the 0.5\% class filter of the official implementation excludes every class that falls below the threshold in the ground truth, where its instances are small or seldom visible, or in the prediction, so that the set of scored classes depends on the prediction under evaluation.
Because the threshold is a share of all labeled pixels, a class that appears in only some of the views is more likely to fall below it in a scene with many views.
Standard panoptic and instance segmentation~\citep{kirillov2019panoptic,lin2014coco} exclude no class on account of its size.

A metric for this task should therefore meet requirements (i) and (ii) and count over the dataset with the input set in place of the image.

\section{View-Consistent Panoptic Quality}
\label{sec:vcpq}

\subsection{Definition}
\label{sec:vcpq_def}

To this end, we propose View-Consistent Panoptic Quality (VC-PQ), which extends PQ from a single image to a set of views.
VC-PQ takes the input set of $N$ views as its unit of evaluation.
Within each set, each pair of a predicted instance $p$ and a ground-truth instance $g$ of the same class receives a pair score $q(p,g) \in [0,1]$.

An instance $x$, whether predicted or ground truth, is regarded as visible in view $i$ when its mask $M_i(x)$ has at least $\eta$ pixels, and its visibility support $V(x)$ is the set of these views, \mbox{$V(x) = \{\, i \mid \lvert M_i(x) \rvert \ge \eta \,\}$}, where $\eta$ is a visibility threshold specified in pixels at the evaluation resolution and set to 100 following PanopticNeRF-360~\citep{fu2022panopticnerf}.

For a predicted instance $p$ and a ground-truth instance $g$, $\mathrm{IoU}_i(p,g)$ is the per-view IoU of Sec.~\ref{sec:pqscene_mismatch}.
Two quantities enter the pair score: the mean IoU over the views in which both are visible, each view counting equally, and a view-consistency factor given by the Jaccard index of their supports,
\begin{equation}
  \overline{\mathrm{IoU}}(p,g) = \frac{1}{\lvert V(p) \cap V(g) \rvert} \smashoperator[r]{\sum_{i \in V(p) \cap V(g)}} \mathrm{IoU}_i(p,g), \qquad \mathrm{VC}(p,g) = \frac{\lvert V(p) \cap V(g) \rvert}{\lvert V(p) \cup V(g) \rvert},
  \label{eq:iou_vc}
\end{equation}
which address requirements (i) and (ii), respectively, with $\overline{\mathrm{IoU}}(p,g)$ taken as zero when $V(p) \cap V(g)$ is empty.
The pair score is the product of the two quantities,
\begin{equation}
  q(p,g) = \overline{\mathrm{IoU}}(p,g)\,\mathrm{VC}(p,g) = \frac{1}{\lvert V(p) \cup V(g) \rvert} \smashoperator[r]{\sum_{i \in V(p) \cap V(g)}} \mathrm{IoU}_i(p,g),
  \label{eq:q}
\end{equation}
that is, the IoU averaged over the union of the two supports, where the views outside their intersection count as zero.

Within each input set and class, predicted and ground-truth instances are matched one-to-one among the pairs with $q(p,g) > 0.5$, the same threshold as in PQ, which makes the matching unique since no instance can exceed this threshold with more than one counterpart (Appendix~\ref{app:definitions}).
An identity split across views lowers the view-consistency factor of every fragment, and with it the pair score, which may fall to 0.5 or below so that the fragment is no longer matched as a TP.
Matched pairs, unmatched predicted instances, and unmatched ground-truth instances are counted over the dataset as TP, FP, and FN, as in PQ.
VC-PQ is then defined as PQ with the IoU of each matched pair replaced by its pair score,
\begin{equation}
  \text{VC-PQ} = \frac{\sum_{(p,g) \in \mathrm{TP}} q(p,g)}{\lvert \mathrm{TP} \rvert + \tfrac{1}{2}\lvert \mathrm{FP} \rvert + \tfrac{1}{2}\lvert \mathrm{FN} \rvert},
  \label{eq:vcpq}
\end{equation}
computed for each class independently and averaged over classes.

\subsection{Relation to \texorpdfstring{PQ$^{\text{scene}}$}{PQ-scene}}
\label{sec:vcpq_gamma}
VC-PQ addresses each of the three properties of Sec.~\ref{sec:pqscene_mismatch} in a way that matches the task.
In contrast to the area weighting of tiling, every view in the union of Eq.~\eqref{eq:q} contributes one term regardless of how large the instances appear in it, so that a missed or spurious appearance lowers the score by one term whatever its area.
In place of per-scene averaging, matching within the input set and counting over the dataset keep the structure of PQ.
The visibility threshold $\eta$, the only size-related criterion in VC-PQ, keeps tiny fragments of noisy annotation or prediction from enlarging the union in Eq.~\eqref{eq:q}.
It acts on individual appearances rather than on classes, so that an instance is excluded from evaluation only when its mask falls below $\eta$ pixels in every view of the set.
Apart from the visibility threshold, VC-PQ therefore differs from PQ only in how a pair is scored, and with a single view, where $q(p,g)$ equals $\mathrm{IoU}_i(p,g)$, it coincides with PQ.

The first of the three properties, the area weighting, can be varied on its own.
Written per view, Eq.~\eqref{eq:iou_concat} is the average of the per-view IoU weighted by $a_i = \lvert M_i(p) \cup M_i(g) \rvert$, the union area of the pair in image $I_i$.
Raising these weights to a power $\gamma \ge 0$ and restricting the sum to the support union of Eq.~\eqref{eq:q} gives a one-parameter family of pair scores,
\begin{equation}
  q_\gamma(p,g) = \smashoperator[r]{\sum_{i \in V(p) \cup V(g)}} w_i \,\mathrm{IoU}_i(p,g), \qquad w_i = \frac{a_i^{\gamma}}{\sum_{j \in V(p) \cup V(g)} a_j^{\gamma}},
  \label{eq:q_gamma}
\end{equation}
with $\mathrm{IoU}_i(p,g)$ taken as zero outside $V(p) \cap V(g)$.
At $\gamma = 0$ every view weighs the same and $q_\gamma$ coincides with $q$ of Eq.~\eqref{eq:q}, and at $\gamma = 1$ the weights are the areas $a_i$.
We write VC-PQ$_\gamma$ for VC-PQ computed with $q_\gamma$ in place of $q$, so that VC-PQ = VC-PQ$_0$ and VC-PQ$_1$ scores every pair by the area weighting of tiling.
Between these ends the unit of evaluation, the counting over the dataset, and the absence of a class filter stay fixed.
Sweeping $\gamma$ therefore shows how much of a method's score rests on the views in which its instances appear large, which Sec.~\ref{sec:gamma_sweep} measures.

\subsection{Diagnostic decomposition}
\label{sec:vcpq_decomp}
The pair score of Eq.~\eqref{eq:q} is the product of two factors, each addressing one of requirements (i) and (ii), so that the score a prediction does not receive can be attributed to the requirement it fails to meet.
Averaging $\mathrm{VC}(p,g)$ over the matched pairs would not show this, since the instances lost to view consistency are precisely those that fail to match.
Instead, we attribute the whole of the score not received, $1 - \text{VC-PQ}$, to the two requirements and to the matching threshold,
\begin{equation}
  \begin{split}
    1 - \text{VC-PQ} = {}& \underbrace{\frac{\sum_{\mathrm{TP}}\bigl(1-\overline{\mathrm{IoU}}\bigr) + \tfrac12\lvert\mathrm{FN}_{\overline{\mathrm{IoU}}}\rvert + \tfrac12\lvert\mathrm{FP}_{\overline{\mathrm{IoU}}}\rvert}{D}}_{\varepsilon_{\overline{\mathrm{IoU}}}} \\
    & + \underbrace{\frac{\sum_{\mathrm{TP}}\overline{\mathrm{IoU}}\,\bigl(1-\mathrm{VC}\bigr) + \tfrac12\lvert\mathrm{FN}_{\mathrm{VC}}\rvert + \tfrac12\lvert\mathrm{FP}_{\mathrm{VC}}\rvert}{D}}_{\varepsilon_{\mathrm{VC}}} + \underbrace{\frac{\tfrac12\lvert\mathrm{FN}_{q}\rvert + \tfrac12\lvert\mathrm{FP}_{q}\rvert}{D}}_{\varepsilon_{q}},
  \end{split}
  \label{eq:decomp}
\end{equation}
where $D$ denotes the denominator of Eq.~\eqref{eq:vcpq}, the sums run over the matched pairs, and $\mathrm{FN}_{\overline{\mathrm{IoU}}}$, $\mathrm{FN}_{\mathrm{VC}}$, and $\mathrm{FN}_{q}$ partition FN as follows, and likewise for FP with the roles of $p$ and $g$ exchanged.

An unmatched instance is classified by three tests applied in turn to its candidates, the predicted instances of its class in the input set: $\overline{\mathrm{IoU}}(p,g) > 0.5$, then $\mathrm{VC}(p,g) > 0.5$, then $q(p,g) > 0.5$.
A ground-truth instance $g$ belongs to $\mathrm{FN}_{\overline{\mathrm{IoU}}}$ if no candidate passes the first test, to $\mathrm{FN}_{\mathrm{VC}}$ if some does but none of those passes the second, and to $\mathrm{FN}_{q}$ if some passes both and fails only at the third.
Since a pair with $q(p,g) > 0.5$ is always matched and both factors of Eq.~\eqref{eq:q} are at most one, every unmatched instance falls in exactly one of the three (Fig.~\ref{fig:decomp_regions}, Appendix~\ref{app:decomposition}).

For a matched pair, the cross term $(1 - \overline{\mathrm{IoU}})(1 - \mathrm{VC})$ of $1 - q$ is counted with the mean IoU in Eq.~\eqref{eq:decomp}, so that $\varepsilon_{\mathrm{VC}}$ is a lower bound on the score lost to view consistency, and $\varepsilon_{q}$ measures the cost of combining the two factors by a product.
The three terms thus separate the errors that occur within a single image from those that arise only across views, since $\varepsilon_{\mathrm{VC}}$ and $\varepsilon_{q}$ vanish when the set contains a single view and $1 - \mathrm{PQ} = \varepsilon_{\overline{\mathrm{IoU}}}$ up to the visibility threshold.
Raising either factor would change the matching itself, so the terms attribute the lost score rather than predict a gain.

\begin{table}[t]
\centering\small
\caption{PQ$^{\text{scene}}$ and VC-PQ of the same predictions for input sets of $N$ views, all classes; ScanNet++ pools the DSLR and iPhone runs. --: out of memory at $N \geq 20$.}
\label{tab:scores}
\begin{tabular}{l l cccc cccc}
\toprule
 & & \multicolumn{4}{c}{PQ$^{\text{scene}}$} & \multicolumn{4}{c}{VC-PQ} \\
\cmidrule(lr){3-6}\cmidrule(lr){7-10}
Dataset & Method & $N{=}5$ & $N{=}10$ & $N{=}20$ & $N{=}40$ & $N{=}5$ & $N{=}10$ & $N{=}20$ & $N{=}40$ \\
\midrule
ScanNet++ & PanSt3R v1 & 43.7 & 45.7 & 48.9 & 50.4 & 22.0 & 19.2 & 17.9 & 17.4 \\
 & PanSt3R v2 & 46.8 & 51.3 & 54.7 & 57.7 & 24.5 & 22.2 & 20.7 & 20.0 \\
 & EPS3D      & 13.2 & 13.6 & -- & -- & 2.4 & 2.0 & -- & -- \\
 & IGGT       & 17.8 & 17.1 & -- & -- & 5.7 & 4.2 & -- & -- \\
\midrule
ScanNetv2 & PanSt3R v2 & 50.1 & 52.5 & 55.2 & 57.4 & 43.7 & 41.5 & 40.6 & 40.6 \\
 & EPS3D      & 21.7 & 21.5 & -- & -- & 9.2 & 8.2 & -- & -- \\
\bottomrule
\end{tabular}
\end{table}

\section{Experiments}
\label{sec:experiments}

\subsection{Evaluation protocol}
\label{sec:eval_protocol}
We evaluate on the validation splits of ScanNet++~\citep{yeshwanth2023scannetpp} and ScanNetv2~\citep{dai2017scannet}.
ScanNet++ provides 50 validation scenes, each captured twice, with a DSLR camera and with an iPhone,\footnote{One iPhone capture is excluded because its ground truth is unusable, leaving 49 iPhone scenes.} and we use its panoptic benchmark of 100 classes, 84 things and 16 stuff.
ScanNetv2 provides 312 validation scenes with 20 classes, 18 things and 2 stuff.

For each scene we sample 120 views with a fixed seed and split them at random into input sets of $N$ views, so that the views of a set are spread over the scene with no continuity from one view to the next (Sec.~\ref{sec:problem_formulation}).
Every method receives the same input sets and predicts a panoptic label map for each view of a set.
We take $N = 10$ as the default, following the multi-view pose evaluation of feed-forward reconstruction on 10 random frames per sequence~\citep{dust3r,wang2025vggt}, and vary $N \in \{5, 10, 20, 40\}$ below and above it in Tables~\ref{tab:scores} and~\ref{tab:decomp}.

Panoptic Lifting computes PQ$^{\text{scene}}$ over the held-out views of each scene, one quarter of its frames, and averages it over the 26 scenes of its benchmark.
To score the same input as VC-PQ, we treat each input set as the views of a scene, as feed-forward reconstruction does, tile it into a single image as in the official implementation~\citep{siddiqui2023panoptic}, and sum the counts of the sets of a scene into one value per scene before averaging over scenes.
For both metrics, void pixels are handled as in PQ~\citep{kirillov2019panoptic}: predicted pixels on ground-truth void are removed from the prediction, and an unmatched predicted instance with more than half of its pixels on void is not counted as FP.
Further details of the protocol are given in Appendix~\ref{app:protocol}.
The evaluation code and the input sets of the protocol are available at \url{https://github.com/Renem24/vcpq}.

\subsection{Evaluating feed-forward methods}
\label{sec:comparison}
We evaluate four publicly available feed-forward methods: PanSt3R~\citep{panst3r} in its two released versions (v1 and v2), EPS3D~\citep{eps3d}, IGGT~\citep{iggt}, and SegVGGT~\citep{segvggt}.
PanSt3R and EPS3D output panoptic labels and are run with their native post-processing.
IGGT outputs class-agnostic instances, which we label with the open-vocabulary model LSeg~\citep{li2022lseg} as EPS3D does.
SegVGGT outputs thing instances, which we turn into panoptic labels with the panoptic inference of Mask2Former~\citep{cheng2022mask2former} and evaluate over its thing classes (Table~\ref{tab:decomp} and Appendix~\ref{app:methods}).
The two metrics score the same predictions of each method, so the comparison between them does not depend on how the methods were trained.

Table~\ref{tab:scores} reports PQ$^{\text{scene}}$ and VC-PQ of the same predictions at all input set sizes.
EPS3D and IGGT score low under both metrics, which we attribute to how they form instances (Appendix~\ref{app:methods}).
PanSt3R v2 scores above v1 under both metrics, and for both versions the two metrics move in opposite directions as $N$ grows.
PQ$^{\text{scene}}$ rises from 46.8 at $N = 5$ to 57.7 at $N = 40$ for v2 on ScanNet++ and from 50.1 to 57.4 on ScanNetv2, whereas VC-PQ falls for every method, from 24.5 to 20.0 and from 43.7 to 40.6 for v2.
Overall, VC-PQ is lower than PQ$^{\text{scene}}$ for every method and every $N$.
Sec.~\ref{sec:gamma_sweep} traces both differences to the three properties of PQ$^{\text{scene}}$.

\begin{figure}[t]
\centering
\includegraphics[width=\textwidth]{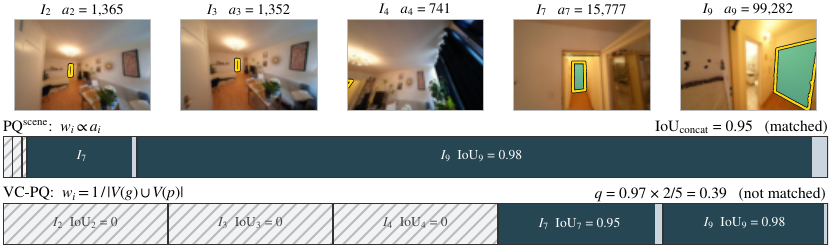}
\caption{A door (PanSt3R v2, ScanNet++ DSLR, $N{=}10$) visible in five views and predicted in the two largest: PQ$^{\text{scene}}$ matches the pair at $\mathrm{IoU}_{\text{concat}} = 0.95$, VC-PQ leaves it unmatched at $q = 0.39$. The union area $a_9$ in $I_9$ is 84\% of the total, so this view alone contributes 0.82 to $\mathrm{IoU}_{\text{concat}}$, whereas VC-PQ counts the door as missed in three of its five views, $\mathrm{VC} = 2/5$.}
\label{fig:case}
\end{figure}
\begin{table}[t]
\centering\small
\caption{VC-PQ with the three properties of PQ$^{\text{scene}}$ switched on in every combination (ScanNet++, $N{=}10$, all classes). $^\dagger$Differs from PQ$^{\text{scene}}$ of Table~\ref{tab:scores} only by the visibility threshold. Right block: Shapley attribution of the difference between the two named columns.}
\label{tab:properties}
\setlength{\tabcolsep}{4.4pt}
\begin{tabular}{l cccccccc @{\hspace{10pt}} cccc}
\toprule
 & VC-PQ & & & & & & & PQ$^{\text{scene}\dagger}$ & \multicolumn{4}{c}{PQ$^{\text{scene}\dagger}$ $-$ VC-PQ, Shapley} \\
\cmidrule(lr){2-9}\cmidrule(lr){10-13}
area weighting & \textcolor{gray!50}{--} & \checkmark & \textcolor{gray!50}{--} & \textcolor{gray!50}{--} & \checkmark & \checkmark & \textcolor{gray!50}{--} & \checkmark &  & \checkmark &  &  \\
per-scene avg. & \textcolor{gray!50}{--} & \textcolor{gray!50}{--} & \checkmark & \textcolor{gray!50}{--} & \checkmark & \textcolor{gray!50}{--} & \checkmark & \checkmark &  &  & \checkmark &  \\
0.5\% filter & \textcolor{gray!50}{--} & \textcolor{gray!50}{--} & \textcolor{gray!50}{--} & \checkmark & \textcolor{gray!50}{--} & \checkmark & \checkmark & \checkmark &  &  &  & \checkmark \\
\midrule
PanSt3R v1 & 19.2 & 25.0 & 24.2 & 26.0 & 29.7 & 35.6 & 37.2 & 45.8 & +26.6 & +7.3 & +7.7 & +11.5 \\
PanSt3R v2 & 22.2 & 28.9 & 25.6 & 32.7 & 31.2 & 42.7 & 42.3 & 51.4 & +29.2 & +7.8 & +6.0 & +15.3 \\
EPS3D & 2.0 & 2.8 & 7.0 & 2.5 & 8.8 & 3.4 & 11.1 & 13.9 & +11.9 & +1.7 & +7.6 & +2.7 \\
IGGT & 4.2 & 6.2 & 7.0 & 6.0 & 9.5 & 8.7 & 12.6 & 17.1 & +12.9 & +3.0 & +5.4 & +4.5 \\
\bottomrule
\end{tabular}
\end{table}

\subsection{Comparison with \texorpdfstring{PQ$^{\text{scene}}$}{PQ-scene}}
\label{sec:gamma_sweep}
\label{sec:perturbation}
Sec.~\ref{sec:pqscene_mismatch} locates the failure of both requirements in the area weighting of tiling, and Fig.~\ref{fig:case} shows one instance of the predictions of Table~\ref{tab:scores} on which the two metrics disagree for this reason.
To isolate the effect of that weighting, we perturb the ground truth by removing an instance from, or changing its ID in, a fixed share of the views in which it is visible, chosen among the smallest, the largest, or random views (Table~\ref{tab:perturb}).
VC-PQ is the same regardless of the mask area in the affected views, whereas PQ$^{\text{scene}}$ falls from 91.4 to 39.3 as those views change from the smallest to the largest.

To locate the difference between the two metrics more precisely, Table~\ref{tab:properties} switches the three properties of PQ$^{\text{scene}}$ on within VC-PQ in every combination over all classes.
With all three on, VC-PQ reproduces PQ$^{\text{scene}}$ to within about 0.3 ($\dagger$), so that the three properties account for the difference between the two metrics.
Switching any one property on raises the score of every method, whichever of the other two properties are already on.
The right block splits the difference between VC-PQ and the score with all three on over the three properties by Shapley attribution~\citep{shapley1953value}, so that the three shares sum to the difference.
For PanSt3R v2 the difference of 29.2 splits into 7.8 from the area weighting, 6.0 from per-scene averaging, and 15.3 from the filter.
For EPS3D and IGGT the largest share comes instead from per-scene averaging, but over thing classes (Table~\ref{tab:properties_things}) that share becomes the smallest for every method, so it comes from stuff classes.
Otherwise the filter takes a large share of the difference for every method, the largest for PanSt3R.

The opposite trends with $N$ arise from two of these properties, the area weighting and per-scene averaging.
Sweeping $\gamma$ for PanSt3R v2 (Fig.~\ref{fig:gamma_N}), the score at $\gamma = 0$ falls from 24.5 at $N = 5$ to 20.0 at $N = 40$ on ScanNet++, whereas at $\gamma = 1$ it stays between 28.8 and 29.2, so the fall of VC-PQ with $N$ comes with the equal weighting of views.
Under per-scene averaging, in contrast, VC-PQ rises with $N$, from 25.3 to 28.2, as PQ$^{\text{scene}}$ does (Appendix~\ref{app:comparison}).

\begin{table}[t]
\centering\footnotesize
\caption{Decomposition of $1 - \text{VC-PQ}$ by Eq.~\eqref{eq:decomp} for input sets of $N$ views, thing classes; VC-PQ and the three terms sum to 100.}
\label{tab:decomp}
\setlength{\tabcolsep}{1.9pt}
\begin{tabular}{l cccc cccc cccc cccc}
\toprule
& \multicolumn{4}{c}{$N{=}5$} & \multicolumn{4}{c}{$N{=}10$} & \multicolumn{4}{c}{$N{=}20$} & \multicolumn{4}{c}{$N{=}40$} \\
\cmidrule(lr){2-5}\cmidrule(lr){6-9}\cmidrule(lr){10-13}\cmidrule(lr){14-17}
Method & VC-PQ & $\varepsilon_{\overline{\mathrm{IoU}}}$ & $\varepsilon_{\mathrm{VC}}$ & $\varepsilon_{q}$ & VC-PQ & $\varepsilon_{\overline{\mathrm{IoU}}}$ & $\varepsilon_{\mathrm{VC}}$ & $\varepsilon_{q}$ & VC-PQ & $\varepsilon_{\overline{\mathrm{IoU}}}$ & $\varepsilon_{\mathrm{VC}}$ & $\varepsilon_{q}$ & VC-PQ & $\varepsilon_{\overline{\mathrm{IoU}}}$ & $\varepsilon_{\mathrm{VC}}$ & $\varepsilon_{q}$ \\
\midrule
\multicolumn{17}{l}{\textit{ScanNet++}} \\
PanSt3R v1 & 21.8 & 73.1 & 4.1 & 1.0 & 18.7 & 74.0 & 4.7 & 2.5 & 17.4 & 74.0 & 4.3 & 4.3 & 16.9 & 73.6 & 3.9 & 5.6 \\
PanSt3R v2 & 24.7 & 68.7 & 5.2 & 1.4 & 22.3 & 67.5 & 7.2 & 3.0 & 20.6 & 66.7 & 7.5 & 5.2 & 19.9 & 66.3 & 7.0 & 6.8 \\
EPS3D & 0.6 & 98.8 & 0.5 & 0.1 & 0.3 & 99.1 & 0.4 & 0.2 & -- & -- & -- & -- & -- & -- & -- & -- \\
IGGT & 4.9 & 93.0 & 1.8 & 0.3 & 3.7 & 93.7 & 1.9 & 0.7 & -- & -- & -- & -- & -- & -- & -- & -- \\
\midrule
\multicolumn{17}{l}{\textit{ScanNetv2}} \\
PanSt3R v2 & 40.6 & 49.4 & 8.0 & 2.0 & 38.1 & 49.7 & 8.4 & 3.7 & 37.1 & 50.0 & 6.9 & 6.0 & 37.1 & 50.1 & 5.8 & 7.1 \\
SegVGGT & 48.7 & 44.0 & 5.8 & 1.5 & 47.0 & 44.2 & 5.9 & 2.9 & 45.8 & 44.1 & 5.4 & 4.7 & 44.9 & 44.5 & 5.0 & 5.6 \\
\bottomrule
\end{tabular}
\end{table}

\subsection{Error decomposition}
\label{sec:error_decomp}
Table~\ref{tab:decomp} applies the decomposition of Eq.~\eqref{eq:decomp} to each method over thing classes, for which an ID across views is defined.
The first term collects the errors within single images and the other two those that arise only across views (Sec.~\ref{sec:vcpq_decomp}), so that the table reads as two shares.
For every method and every $N$ the largest term is $\varepsilon_{\overline{\mathrm{IoU}}}$, above 90 for EPS3D and IGGT and 44 to 74 for the others, and it changes little with $N$.
Most of the score is thus lost within single images.
The two terms across views are what VC-PQ is designed to penalize.
$\varepsilon_{\mathrm{VC}}$ collects the instances whose views disagree severely, with the prediction and the ground truth sharing at most half of the views in which either appears.
$\varepsilon_{q}$ collects the instances whose views disagree and whose masks are inaccurate, neither enough to fail on its own, so that the product of the two falls below the matching threshold.
Together, the two terms measure how much of the score a method loses to the multi-view setting.

As $N$ grows, $\varepsilon_{\mathrm{VC}}$ changes little and in no consistent direction, so the severe disagreements across views remain a roughly constant share of the score.
$\varepsilon_{q}$ instead rises for every method and dataset, from 1.4 to 6.8 for PanSt3R v2 on ScanNet++ and from 2.0 to 7.1 on ScanNetv2, so more instances fall below the matching threshold on the combination of the two factors.
Since $\varepsilon_{\overline{\mathrm{IoU}}}$ also changes little, the fall of VC-PQ with $N$ is accounted for by this term.
Appendix~\ref{app:decomposition} traces the rise to the predictions rather than to the metric, since an instance is matched with the same probability at every $N$ given the fraction of the input set in which it is visible, and larger sets hold more instances that are visible in only a few views (Figs.~\ref{fig:coverage} to \ref{fig:coverage_dist}).

The decomposition thus shows in which respect each method loses its score, and hence in which direction it should improve.
PanSt3R v2 improves on v1 in $\varepsilon_{\overline{\mathrm{IoU}}}$, whereas its $\varepsilon_{\mathrm{VC}}$ and $\varepsilon_{q}$ are larger.
On ScanNetv2 SegVGGT is ahead of PanSt3R v2 in all three terms, most in $\varepsilon_{\overline{\mathrm{IoU}}}$.
EPS3D and IGGT lose nearly all of their score before view consistency enters.

\section{Conclusion}
\label{sec:conclusion}
We have formulated multi-view panoptic segmentation on the input views and proposed VC-PQ, which counts every view in which an instance is visible equally and penalizes a prediction that is not visible in the same views as its ground truth.
On recent feed-forward methods the two metrics move in opposite directions as the input set grows, which varying the area weighting through $\gamma$ traces to the equal weighting of views.
The decomposition separates the score lost within single images from the part that the multi-view setting costs.
We hope that VC-PQ will be reported alongside PQ$^{\text{scene}}$, and that the decomposition serves to locate where a method loses its score and thus to guide its development in the right direction.

VC-PQ is defined for a static scene and a fixed set of input views, and the evaluation covers indoor scans of two datasets and four methods.
In indoor scans the ground-truth masks are projected from 3D annotations and are therefore noisy, and VC-PQ, which counts every view equally, can be sensitive to such noise in the mask of an individual view (Table~\ref{tab:erosion}).
The visibility threshold $\eta$ and a slight area weighting through $\gamma$, which shifts weight toward the views in which an instance appears large, are the two means the metric provides against such noise.

\bibliography{main}
\bibliographystyle{iclr2027_conference}

\clearpage
\appendix
\section*{Appendix}
\suppressfloats[t]
\renewcommand{\thetable}{A\arabic{table}}\setcounter{table}{0}
\renewcommand{\thefigure}{A\arabic{figure}}\setcounter{figure}{0}
\renewcommand{\theHtable}{A\arabic{table}}\renewcommand{\theHfigure}{A\arabic{figure}}

\section{Evaluation protocol details}
\label{app:protocol}

ScanNet++ has 50 validation scenes, each captured with a DSLR camera at $920 \times 613$ and with an iPhone at $960 \times 720$, and ScanNetv2 has 312 validation scenes at $1296 \times 968$.
From every capture we sample 120 views with seed 42, shuffle them, and slice the sequence into input sets of $N$ consecutive elements, which gives 24, 12, 6, and 3 sets per scene for $N = 5, 10, 20, 40$.
Within a set the views are ordered by their capture index.
At $N = 10$ a set holds on average 51.2 ground-truth instances above the visibility threshold on the DSLR captures, 44.0 on the iPhone captures, and 13.9 on ScanNetv2.
For comparison, Panoptic Lifting reports PQ$^{\text{scene}}$ on 26 scenes of HyperSim, ScanNet, and Replica (6, 12, and 8 scenes), with one quarter of the frames of each scene held out as test views~\citep{siddiqui2023panoptic}.

PanSt3R v1 assigns pixels to queries by simulated annealing, which is run once per input set without a fixed seed, so a second run of the same set can give a different label map.
Between two such runs on the same input sets, PQ$^{\text{scene}}$ differs by at most 0.6 and VC-PQ by at most 0.1 at $N = 10$ on either camera.
PanSt3R v2 is run with its released two-pass assignment and class-score threshold of 0.1.
EPS3D receives the full frames of a set and does not run at $N \geq 20$ on a 24\,GB GPU, nor does the clustering of IGGT.
PanSt3R and EPS3D are trained on ScanNet++, EPS3D also on ScanNet, and IGGT on ScanNet++ among other datasets. The training split of EPS3D on ScanNet++ is not specified, so an overlap with the validation scenes cannot be excluded, and the comparison across methods does not control for training data.
IGGT outputs an instance embedding per pixel and clusters it over the views of a set. Each cluster is given the class whose text embedding is closest to the average of the LSeg~\citep{li2022lseg} pixel features over the cluster mask, with the same prompts as EPS3D.
SegVGGT is run from its released checkpoint on ScanNetv2, and its instance masks and class scores are turned into a panoptic label map with the panoptic inference of Mask2Former~\citep{cheng2022mask2former}, and pixels that no mask claims are void.
Its masks are binarized at 0.4, masks with fewer than 200 pixels are dropped, each query takes its highest-scoring class, its class score is multiplied by the mean mask probability (obj\_normalization), and no threshold is applied to the class score.
Table~\ref{tab:otherfurniture} lists the 41 prompts that the open-vocabulary methods receive on ScanNetv2, namely the names of the other 19 classes and, in place of otherfurniture, the 22 ScanNet200 categories that the official label mapping assigns to it.

\begin{table}[htp]
\centering\small
\caption{The 41 prompts given to the open-vocabulary methods on ScanNetv2, in alphabetical order within each group (read down each column). The first group are the names of the other 19 classes, each scored as its own class. The second group are the 22 ScanNet200 categories~\citep{rozenberszki2022language} that the official label mapping assigns to otherfurniture, all scored as otherfurniture.}
\label{tab:otherfurniture}
\begin{tabular}{llll}
\toprule
\multicolumn{4}{l}{\textit{Scored as its own class (19)}} \\
\midrule
bathtub & counter & picture & table \\
bed & curtain & refrigerator & toilet \\
bookshelf & desk & shower curtain & wall \\
cabinet & door & sink & window \\
chair & floor & sofa &  \\
\midrule
\multicolumn{4}{l}{\textit{Scored as otherfurniture (22)}} \\
\midrule
bench & keyboard piano & radiator & tv stand \\
closet & ladder & recycling bin & wardrobe \\
clothes dryer & music stand & seat & washing machine \\
crate & ottoman & stand & water cooler \\
furniture & piano & trash bin &  \\
ironing board & rack & trash can &  \\
\bottomrule
\end{tabular}
\end{table}

Since the DSLR and iPhone runs of ScanNet++ capture the same scenes, confidence intervals are obtained by a bootstrap over the 50 physical scenes that keeps both captures of a drawn scene together, and on ScanNetv2 it draws the 312 scenes.
For each of $B = 10{,}000$ draws both metrics are recomputed from the per-set counts with the aggregation of Sec.~\ref{sec:eval_protocol}, and the interval is the 95\% BCa interval~\citep{efron1987better}, whose acceleration constant is estimated by a leave-one-scene-out jackknife.
The tables of the paper give point estimates, and Table~\ref{tab:confidence} gives the intervals at $N = 10$.

\section{Additional definitions}
\label{app:definitions}

PQ factors into a segmentation quality and a recognition quality, $\mathrm{PQ} = \mathrm{SQ} \times \mathrm{RQ}$, with SQ the mean IoU over the matched pairs and RQ the F1 score of the matching~\citep{kirillov2019panoptic}.
VC-PQ factors in the same way with the pair score in place of the IoU, \[\text{VC-SQ} = \frac{\sum_{(p,g) \in \mathrm{TP}} q(p,g)}{\lvert \mathrm{TP} \rvert}, \qquad \text{VC-RQ} = \frac{\lvert \mathrm{TP} \rvert}{\lvert \mathrm{TP} \rvert + \tfrac{1}{2}\lvert \mathrm{FP} \rvert + \tfrac{1}{2}\lvert \mathrm{FN} \rvert},\] so that $\text{VC-PQ} = \text{VC-SQ} \times \text{VC-RQ}$ for each class.
As for PQ, the product holds per class and not for the class averages.
Table~\ref{tab:sqrq} in Appendix~\ref{app:sqrq} gives the two factors at every $N$.

The matching of Sec.~\ref{sec:vcpq} is unique because no instance can have a pair score above 0.5 with two counterparts.
For a predicted instance $p$ and two ground-truth instances $g$ and $g'$ of the same class, the masks $M_i(g)$ and $M_i(g')$ are disjoint in every view, so $\mathrm{IoU}_i(p,g) + \mathrm{IoU}_i(p,g') \le 1$ for every $i$.
Summing over the views of $V(p)$ and dividing by $\lvert V(p) \cup V(g) \rvert \ge \lvert V(p) \rvert$ gives $q(p,g) + q(p,g') \le 1$, so at most one of the two exceeds 0.5.
The same argument applies to two predicted instances and one ground-truth instance.

The 0.5\% class filter of PQ$^{\text{scene}}$ (Sec.~\ref{sec:pqscene_mismatch}) is not described in the Panoptic Lifting paper. It is \texttt{get\_non\_robust\_classes\_for\_image} in the official code, applied with a threshold of 0.005 after ground-truth void pixels are removed.
The terms of Eq.~\eqref{eq:decomp} are computed for each class and averaged over classes as VC-PQ is, which preserves the identity.
The identity requires the score of Eq.~\eqref{eq:q}, the product of the two quantities of Eq.~\eqref{eq:iou_vc}, and does not hold for $q_\gamma$ of Eq.~\eqref{eq:q_gamma} with $\gamma > 0$.

\FloatBarrier
\section{Evaluating feed-forward methods, additional results}
\label{app:methods}

Table~\ref{tab:per_camera} gives the scores of Table~\ref{tab:scores} at $N = 10$ for each camera of ScanNet++, and Table~\ref{tab:confidence} the confidence intervals of the pooled values.

\begin{table}[htp]
\centering\small
\caption{PQ$^{\text{scene}}$ and VC-PQ on ScanNet++ at $N{=}10$ for each camera, aggregated as in Table~\ref{tab:scores}. The DSLR and iPhone runs capture the same 50 scenes, and the iPhone values exclude the scene without usable ground truth (Sec.~\ref{sec:eval_protocol}).}
\label{tab:per_camera}
\begin{tabular}{l cc cc}
\toprule
& \multicolumn{2}{c}{DSLR} & \multicolumn{2}{c}{iPhone} \\
\cmidrule(lr){2-3}\cmidrule(lr){4-5}
Method & PQ$^{\text{scene}}$ & VC-PQ & PQ$^{\text{scene}}$ & VC-PQ \\
\midrule
PanSt3R v1 & 48.8 & 20.7 & 42.5 & 17.5 \\
PanSt3R v2 & 54.1 & 22.1 & 48.5 & 22.5 \\
EPS3D      & 13.2 & 1.9  & 14.1 & 2.1 \\
IGGT       & 16.6 & 3.8  & 17.7 & 4.7 \\
\bottomrule
\end{tabular}
\end{table}

\begin{table}[htp]
\centering\small
\caption{95\% BCa confidence intervals~\citep{efron1987better} for the scores of Table~\ref{tab:scores} at $N{=}10$. On ScanNet++ the bootstrap draws the 50 physical scenes with replacement and keeps the DSLR and iPhone runs of a drawn scene together; on ScanNetv2 it draws the 312 scenes. $^{*}$The bound lies at the most extreme of the $B{=}10{,}000$ draws.}
\label{tab:confidence}
\begin{tabular}{l cc cc}
\toprule
& \multicolumn{2}{c}{ScanNet++} & \multicolumn{2}{c}{ScanNetv2} \\
\cmidrule(lr){2-3}\cmidrule(lr){4-5}
Method & PQ$^{\text{scene}}$ & VC-PQ & PQ$^{\text{scene}}$ & VC-PQ \\
\midrule
PanSt3R v1 & 45.7 [43.3, 48.1] & 19.2 [17.2, 21.2] & -- & -- \\
PanSt3R v2 & 51.3 [48.9, 53.8] & 22.2 [20.5, 24.1] & 52.5 [51.0, 54.0] & 41.5 [39.8, 43.3] \\
EPS3D      & 13.6 [12.3, 15.2] & 2.0 [1.8, 2.1]$^{*}$ & 21.5 [20.8, 22.2] & 8.2 [8.0, 8.4] \\
IGGT       & 17.1 [15.1, 19.7] & 4.2 [3.6, 4.8] & -- & -- \\
\bottomrule
\end{tabular}
\end{table}

The aggregation is that of Table~\ref{tab:scores} applied to the runs of one camera, and the iPhone values are over the 49 scenes with usable ground truth.
Table~\ref{tab:things} splits the scores at $N = 10$ into thing and stuff classes and adds SegVGGT on ScanNetv2.

\begin{table}[htp]
\centering\small
\caption{The scores of Table~\ref{tab:scores} at $N{=}10$ over all classes, thing classes, and stuff classes. The identity of an instance across views is defined only for things, since each stuff class forms a single instance, so that on stuff the view-consistency factor measures only the mismatch between the sets of views in which a class appears. SegVGGT predicts instances only and enters in the thing columns.}
\label{tab:things}
\begin{tabular}{l ccc ccc}
\toprule
& \multicolumn{3}{c}{PQ$^{\text{scene}}$} & \multicolumn{3}{c}{VC-PQ} \\
\cmidrule(lr){2-4}\cmidrule(lr){5-7}
Method & all & things & stuff & all & things & stuff \\
\midrule
\multicolumn{7}{l}{\textit{ScanNet++}} \\
PanSt3R v1 & 45.7 & 42.1 & 55.6 & 19.2 & 18.7 & 21.7 \\
PanSt3R v2 & 51.3 & 48.4 & 59.4 & 22.2 & 22.3 & 21.8 \\
EPS3D      & 13.6 & 1.3 & 45.5 & 2.0 & 0.3 & 10.7 \\
IGGT       & 17.1 & 11.8 & 32.5 & 4.2 & 3.7 & 6.7 \\
\midrule
\multicolumn{7}{l}{\textit{ScanNetv2}} \\
PanSt3R v2 & 52.5 & 44.1 & 80.6 & 41.5 & 38.1 & 71.5 \\
EPS3D      & 21.5 & 2.8 & 80.9 & 8.2 & 1.4 & 69.3 \\
SegVGGT    & -- & 46.3 & -- & -- & 47.0 & -- \\
\bottomrule
\end{tabular}
\end{table}

Thing and stuff values of PQ$^{\text{scene}}$ apply the 0.5\% filter over all classes, with the labeled pixels of the set as denominator, and then average the classes of the subset. A scene run without any class of the subset is left out of that column.
Table~\ref{tab:fp_nature} gives, for the thing classes of ScanNet++ at $N = 10$, the counts behind the low scores of EPS3D and IGGT in Sec.~\ref{sec:comparison}.

\begin{table}[htp]
\centering\small\setlength{\tabcolsep}{4pt}
\caption{Predicted instances of thing classes on ScanNet++ ($N{=}10$, DSLR and iPhone). Counts of matched and unmatched instances under VC-PQ, the number of predicted instances per ground-truth instance (both above the visibility threshold), VC-PQ with all classes, things and stuff, merged, the share of predicted instances whose class agrees with the ground-truth instance they overlap most, and the share of ground-truth instances split over two or more predicted instances.}
\label{tab:fp_nature}
\begin{tabular}{l rrr c c c c c}
\toprule
Method & $\lvert\mathrm{TP}\rvert$ & $\lvert\mathrm{FP}\rvert$ & $\lvert\mathrm{FN}\rvert$ & \begin{tabular}[b]{@{}c@{}}pred.\\/ GT\end{tabular} & \begin{tabular}[b]{@{}c@{}}class-agnostic\\VC-PQ\end{tabular} & \begin{tabular}[b]{@{}c@{}}class\\agrees (\%)\end{tabular} & \begin{tabular}[b]{@{}c@{}}GT\\split (\%)\end{tabular} \\
\midrule
PanSt3R v1 & 11,044 & 11,905 & 39,334 & 0.46 & 29.1 & 56 & 21 \\
PanSt3R v2 & 15,351 & 21,625 & 35,027 & 0.73 & 31.9 & 52 & 23 \\
EPS3D & 346 & 38,581 & 50,032 & 0.77 & 4.5 & 22 & 49 \\
IGGT & 2,756 & 18,758 & 47,622 & 0.43 & 14.0 & 31 & 17 \\
\bottomrule
\end{tabular}
\end{table}

EPS3D splits half of the ground-truth instances over two or more of its predictions, and only about a fifth of its predictions carry the class of the instance they overlap most.
IGGT predicts fewer than half as many instances as the ground truth holds, and scoring with all classes merged raises its VC-PQ from 4.2 to 14.0.
The first three columns are the matched and unmatched instances under VC-PQ, and the ratio of predicted to ground-truth instances is $(\lvert\mathrm{TP}\rvert + \lvert\mathrm{FP}\rvert) / (\lvert\mathrm{TP}\rvert + \lvert\mathrm{FN}\rvert)$.
The class-agnostic VC-PQ merges all classes, things and stuff, into one before matching.
For each predicted instance we find the ground-truth instance it overlaps most, with pixels summed over the views of the set and void excluded, and count whether the two classes agree. A prediction that overlaps only void counts as a disagreement.
A ground-truth instance counts as split when two or more predicted instances each cover at least 10\% of its pixels.
The last two columns are counted on the label maps without the visibility threshold, and the two cameras are averaged with weights given by their numbers of input sets, 600 and 588.

\clearpage
\section{Comparison with \texorpdfstring{PQ$^{\text{scene}}$}{PQ-scene}, additional results}
\label{app:comparison}

Table~\ref{tab:error_rates} counts, on the runs of Table~\ref{tab:perturb}, the appearances that the four methods miss and those they cover with an instance other than the one that covers the object in most views of the same input set, as a share of all appearances, the same denominator as the perturbation rates of Table~\ref{tab:perturb}.

\begin{table}[htp]
\centering\small
\caption{Errors of the predictions on the runs of Table~\ref{tab:perturb} (ScanNet++ DSLR, $N{=}10$, thing classes), counted over the same appearances. An appearance is covered when a predicted instance overlaps it with an IoU above 0.5 in that view. The first column is the share of appearances that are not covered, and the second the share of appearances covered by a predicted instance other than the one that covers the instance in the largest number of views of the same input set.}
\label{tab:error_rates}
\begin{tabular}{l cc}
\toprule
Method & not covered (\%) & covered by another ID (\%) \\
\midrule
PanSt3R v1 & 63.9 & 0.4 \\
PanSt3R v2 & 54.8 & 1.1 \\
EPS3D      & 92.0 & 0.6 \\
IGGT       & 80.7 & 1.2 \\
\bottomrule
\end{tabular}
\end{table}

\begin{table}[htp]
\centering\small
\caption{Ground truth of ScanNet++ DSLR ($N{=}10$, thing classes) perturbed at one operating point and scored against the original, so that the unperturbed label maps score 100. In each group the same number of appearances, 28\% of all, is removed (missed appearance) or given a new ID (identity split), and the three variants differ only in which views are chosen: the appearances of smallest area first, of largest area first, or at random (mean of three seeds). Per-image PQ scores each view on its own.}
\label{tab:perturb}
\begin{tabular}{ll cccc}
\toprule
Perturbation & Variant & PQ$^{\text{scene}}$ & VC-PQ & VC-PQ$_{1}$ ($\gamma{=}1$) & per-image PQ \\
\midrule
none & -- & 100.0 & 100.0 & 100.0 & 100.0 \\
\midrule
missed appearance & smallest area first & 91.4 & 80.9 & 92.9 & 86.7 \\
 & largest area first & 39.3 & 80.9 & 50.7 & 86.7 \\
 & random & 72.9 & 80.9 & 77.2 & 86.7 \\
\midrule
identity split & smallest area first & 65.2 & 63.2 & 72.3 & 100.0 \\
 & largest area first & 51.6 & 63.2 & 61.9 & 100.0 \\
 & random & 55.8 & 63.2 & 65.9 & 100.0 \\
\bottomrule
\end{tabular}
\end{table}

Table~\ref{tab:erosion} gives the erosion perturbation of Sec.~\ref{sec:perturbation}.

\begin{table}[htp]
\centering\small
\caption{Scores of the ground truth of Table~\ref{tab:perturb} with every thing mask eroded by $r$ pixels, scored against the unperturbed ground truth. Eroded pixels become void, and the image border is not eroded. The last row is the share of appearances that fall below the visibility threshold after erosion.}
\label{tab:erosion}
\begin{tabular}{l cccc}
\toprule
erosion radius $r$ (px) & 1 & 2 & 4 & 8 \\
\midrule
PQ$^{\text{scene}}$ & 85.9 & 77.0 & 63.3 & 45.2 \\
VC-PQ & 61.3 & 40.2 & 20.3 & 8.7 \\
VC-PQ$_1$ ($\gamma{=}1$) & 70.3 & 51.3 & 30.0 & 15.2 \\
per-image PQ & 61.7 & 41.9 & 24.3 & 12.3 \\
appearances below $\eta$ (\%) & 13.2 & 23.4 & 38.1 & 55.0 \\
\bottomrule
\end{tabular}
\end{table}

Every thing mask is eroded by $r$ pixels with a $3 \times 3$ structuring element, the eroded pixels become void, and the image border is not eroded.
The last row is the share of appearances that fall below the visibility threshold after erosion and thus leave the support of the perturbed instance.
Table~\ref{tab:aggregation_n} gives the two aggregations of Sec.~\ref{sec:gamma_sweep} for PanSt3R v2 at every $N$.

\begin{table}[htp]
\centering\small
\caption{PQ$^{\text{scene}}$ and VC-PQ of PanSt3R v2 on ScanNet++ for input sets of $N$ views under two aggregations. Per-scene averages the class-averaged value of each scene run, and pooled sums the counts of all input sets before the class average. The first and fourth rows are the values of Table~\ref{tab:scores}.}
\label{tab:aggregation_n}
\begin{tabular}{l c l cccc}
\toprule
Score & 0.5\% filter & aggregation & $N{=}5$ & $N{=}10$ & $N{=}20$ & $N{=}40$ \\
\midrule
PQ$^{\text{scene}}$ & on  & per-scene & 46.8 & 51.3 & 54.7 & 57.7 \\
PQ$^{\text{scene}}$ & on  & pooled    & 41.3 & 42.7 & 40.9 & 44.2 \\
VC-PQ               & off & per-scene & 25.3 & 25.6 & 26.9 & 28.2 \\
VC-PQ               & off & pooled    & 24.5 & 22.2 & 20.7 & 20.0 \\
PQ$^{\text{scene}}$ & off & pooled    & 28.6 & 28.6 & 28.6 & 29.1 \\
\bottomrule
\end{tabular}
\end{table}

Per-scene averages the class-averaged value of each scene run, and pooled sums the counts of all sets before the class average.
The first and fourth rows are the values of Table~\ref{tab:scores}, the last row is PQ$^{\text{scene}}$ without the 0.5\% filter under pooled counts, and the per-scene rows average the 99 scene runs.
Table~\ref{tab:eta} repeats VC-PQ at $N = 10$ for $\eta = 50$, 100, and 200 pixels, aggregated as in Table~\ref{tab:scores}.

\begin{table}[htp]
\centering\small
\caption{VC-PQ on ScanNet++ at $N{=}10$ for three values of the visibility threshold $\eta$, aggregated as in Table~\ref{tab:scores}. The last row is the share of ground-truth instances whose mask has fewer than $\eta$ pixels in every view of the set and that are therefore excluded from evaluation.}
\label{tab:eta}
\begin{tabular}{l ccc}
\toprule
& $\eta{=}50$ & $\eta{=}100$ & $\eta{=}200$ \\
\midrule
PanSt3R v1 & 18.5 & 19.2 & 20.2 \\
PanSt3R v2 & 21.5 & 22.2 & 23.4 \\
EPS3D      & 2.0  & 2.0  & 2.1 \\
IGGT       & 4.0  & 4.2  & 4.5 \\
\midrule
ground-truth instances excluded (\%) & 0.0 & 2.5 & 6.9 \\
\bottomrule
\end{tabular}
\end{table}

\begin{figure}[htp]
\centering
\includegraphics[width=\textwidth]{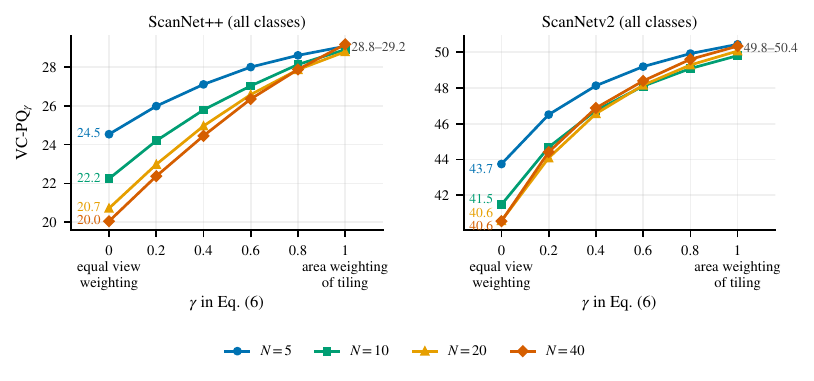}
\caption{VC-PQ$_\gamma$ of PanSt3R v2 against $\gamma$ in Eq.~\eqref{eq:q_gamma}, one curve per input set size $N$, over all classes. The predictions are identical along each curve; only the weighting of views changes. $\gamma=0$ is VC-PQ and $\gamma=1$ applies the area weighting of tiling.}
\label{fig:gamma_N}
\end{figure}

Its last row is the share of ground-truth instances, over thing and stuff classes of both cameras, whose mask falls below $\eta$ in every view of the set and which are therefore excluded from evaluation.
At $\eta = 50$ no instance is excluded, because appearances below 50 pixels are already removed when the label maps are prepared.
Fig.~\ref{fig:gamma_N} sweeps $\gamma$ of Eq.~\eqref{eq:q_gamma} for PanSt3R v2 with one curve per $N$ on both datasets.

Table~\ref{tab:properties_things} repeats Table~\ref{tab:properties} over thing classes, the classes over which Table~\ref{tab:decomp} decomposes the score, so that the two tables can be read on the same classes.
Over thing classes the share of per-scene averaging is the smallest of the three for every method, 20, 12, 24, and 22\% of the difference for PanSt3R v1, PanSt3R v2, EPS3D, and IGGT, whereas over all classes (Table~\ref{tab:properties}) it is the largest for EPS3D and IGGT.

\begin{table}[htp]
\centering\small
\caption{Table~\ref{tab:properties} restricted to thing classes. VC-PQ with the three properties of PQ$^{\text{scene}}$ switched on in every combination, on ScanNet++ at $N{=}10$. $^\dagger$With all three switched on, the score reproduces PQ$^{\text{scene}}$ to within about 0.4 over thing classes of Table~\ref{tab:things}; unlike PQ$^{\text{scene}}$, it applies the visibility threshold. The right block splits the difference between the two named columns over the three properties by Shapley attribution.}
\label{tab:properties_things}
\setlength{\tabcolsep}{4.4pt}
\begin{tabular}{l cccccccc @{\hspace{10pt}} cccc}
\toprule
 & VC-PQ & & & & & & & PQ$^{\text{scene}\dagger}$ & \multicolumn{4}{c}{PQ$^{\text{scene}\dagger}$ $-$ VC-PQ, Shapley} \\
\cmidrule(lr){2-9}\cmidrule(lr){10-13}
area weighting & \textcolor{gray!50}{--} & \checkmark & \textcolor{gray!50}{--} & \textcolor{gray!50}{--} & \checkmark & \checkmark & \textcolor{gray!50}{--} & \checkmark &  & \checkmark &  &  \\
per-scene avg. & \textcolor{gray!50}{--} & \textcolor{gray!50}{--} & \checkmark & \textcolor{gray!50}{--} & \checkmark & \textcolor{gray!50}{--} & \checkmark & \checkmark &  &  & \checkmark &  \\
0.5\% filter & \textcolor{gray!50}{--} & \textcolor{gray!50}{--} & \textcolor{gray!50}{--} & \checkmark & \textcolor{gray!50}{--} & \checkmark & \checkmark & \checkmark &  &  &  & \checkmark \\
\midrule
PanSt3R v1 & 18.7 & 24.5 & 20.8 & 25.2 & 26.1 & 35.0 & 33.5 & 42.2 & +23.5 & +7.3 & +4.8 & +11.4 \\
PanSt3R v2 & 22.3 & 28.9 & 23.1 & 33.1 & 28.3 & 42.6 & 39.8 & 48.5 & +26.1 & +7.5 & +3.2 & +15.4 \\
EPS3D & 0.3 & 0.7 & 0.4 & 0.5 & 0.8 & 1.1 & 0.8 & 1.7 & +1.4 & +0.6 & +0.3 & +0.5 \\
IGGT & 3.7 & 5.7 & 4.0 & 5.3 & 6.1 & 8.0 & 7.5 & 11.8 & +8.1 & +2.9 & +1.8 & +3.4 \\
\bottomrule
\end{tabular}
\end{table}

\begin{figure}[htp]
\centering
\includegraphics[width=\textwidth]{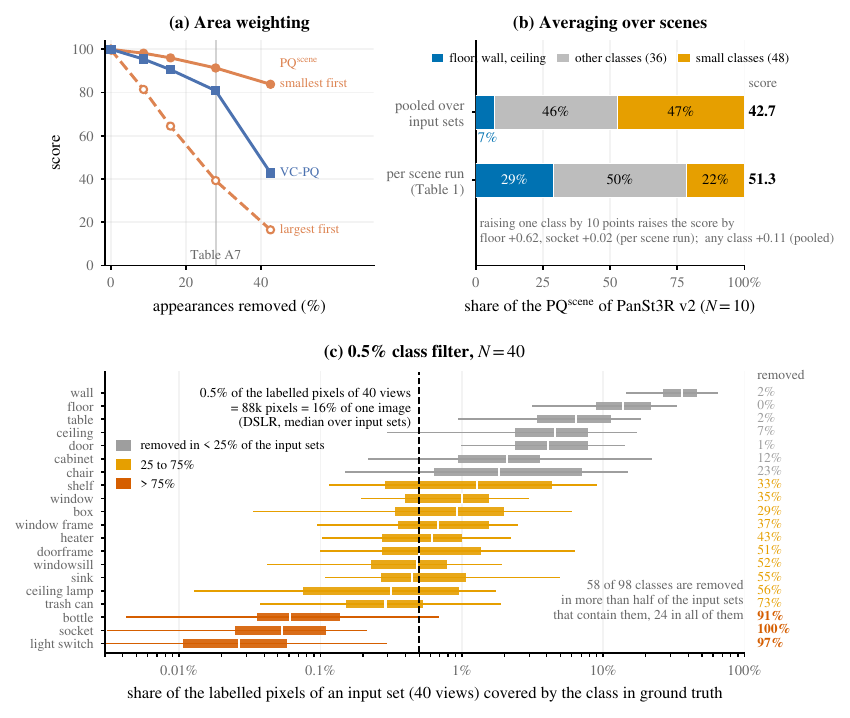}
\caption{The three properties of PQ$^{\text{scene}}$ of Sec.~\ref{sec:pqscene_mismatch} on real data. (a)~Area weighting: ground truth of the ScanNet++ DSLR runs ($N{=}10$, thing classes) with the same number of appearances removed, in the views in which an instance appears smallest or in those in which it appears largest, scored against the unperturbed ground truth; the vertical line marks the operating point of Table~\ref{tab:perturb}. (b)~Averaging over scenes: the counts of PanSt3R v2 on ScanNet++ ($N{=}10$, all classes, 0.5\% filter on), averaged over classes after pooling all input sets and after averaging per scene run (Table~\ref{tab:scores}), split into the shares of floor, wall, and ceiling, of the small classes (median share of the labelled pixels of an input set below 0.5\% at $N{=}10$), and of the other classes. (c)~The 0.5\% class filter at $N{=}40$: for the 20 classes present in the most input sets, the share of the labelled pixels of an input set that the class covers in the ground truth (box: quartiles, line: 5th to 95th percentile), and on the right the share of the input sets containing the class in which the filter removes it.}
\label{fig:pqscene_props}
\end{figure}

The eight combinations and the Shapley attribution are defined as in Table~\ref{tab:properties}, and the row with all three properties switched on differs from PQ$^{\text{scene}}$ over thing classes (Table~\ref{tab:things}) only by the visibility threshold.
Fig.~\ref{fig:pqscene_props} shows the three properties of PQ$^{\text{scene}}$ of Sec.~\ref{sec:pqscene_mismatch} on real data.

In~(a), PQ$^{\text{scene}}$ at the operating point of Table~\ref{tab:perturb} is 91.4 when the smallest appearances are removed and 39.3 when the largest are, whereas VC-PQ is 80.9 in both cases.
In~(b), the same counts give 42.7 when pooled over all input sets and 51.3 when averaged per scene run, and floor, wall, and ceiling account for 7\% of the first score and 29\% of the second.
A per-scene average counts a class once in every scene run in which it is scored, so raising the PQ of floor by 10 points raises the score by 0.62 and that of socket by 0.02, whereas under pooling any class raises it by 0.11.
In~(c), 0.5\% of the labelled pixels of an input set of 40 views is about 88{,}000 pixels (median) on the DSLR captures, 16\% of one image, and over both cameras 58 of the 98 ground-truth classes are removed in more than half of the input sets that contain them.
Table~\ref{tab:perturb} gives the perturbation of Sec.~\ref{sec:perturbation} at its operating point, with the three variants of each perturbation and the four scores.
We take the ground-truth label maps of the ScanNet++ DSLR runs at $N = 10$, restricted to thing classes, apply a perturbation, and score the result against the original, so that the unperturbed maps score 100 under every metric.
An appearance is a pair of an instance and a view in which it is visible, and the two perturbations act on appearances.
A missed appearance removes the instance from the view, and an identity split gives its pixels in that view a new ID of the same class.
Both act on the same number of appearances per instance, about 30\% of the views in which it is visible, which amounts to 28\% of all appearances at the operating point.
They differ only in which views are chosen, namely the views in which the instance appears smallest, those in which it appears largest, or views drawn at random.
The operating point lies below the share of appearances that PanSt3R v2 misses on the same runs (55\%) and far above the share of appearances it covers with an ID other than the one it gives the instance in most views of the same input set (1.1\%).
In addition to the two metrics, the table reports VC-PQ$_1$, which applies the area weighting alone, and per-image PQ, which scores each view independently and is therefore unaffected by a change of ID across views.
The VC-PQ column is identical across the three variants, as predicted before the run, and VC-PQ$_1$ follows PQ$^{\text{scene}}$, which places the difference between the two metrics on the area weighting rather than on the counting or the visibility threshold.

The instance of Fig.~\ref{fig:case} is taken from the ScanNet++ DSLR runs of PanSt3R v2 at $N = 10$, scored by both metrics without the 0.5\% filter ($\eta = 100$, $\gamma = 0$).
It was chosen by hand among the 971 thing instances of this cell that PQ$^{\text{scene}}$ matches and VC-PQ leaves unmatched in the view-consistency test, as a legible example of the area difference, and it is not the highest-ranked case under either of the two orderings we tried.

\FloatBarrier
\section{Error decomposition, additional results}
\label{app:decomposition}

\begin{figure}[htp]
\centering
\includegraphics[width=\textwidth]{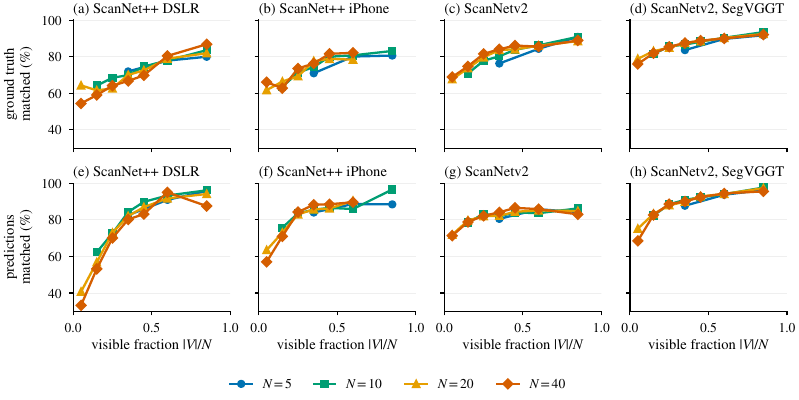}
\caption{Share of matched candidates as a function of the fraction $\lvert V \rvert / N$ of the input set in which the instance is visible, for $\gamma = 0$ (VC-PQ), one curve per $N$, over thing classes visible in at least two views. Top: ground-truth instances, bottom: predicted instances. (a)--(c) and (e)--(g) are PanSt3R v2 on the three captures, (d) and (h) SegVGGT on ScanNetv2. A candidate is an instance with a same-class counterpart at $\overline{\mathrm{IoU}}(p,g) > 0.5$, and bins with fewer than 30 instances are omitted. The fraction is binned, with bin edges at 0.1, 0.2, 0.3, 0.4, 0.5, and 0.7, and each marker sits at the centre of its bin.}
\label{fig:coverage}
\end{figure}

\begin{figure}[htp]
\centering
\includegraphics[width=\textwidth]{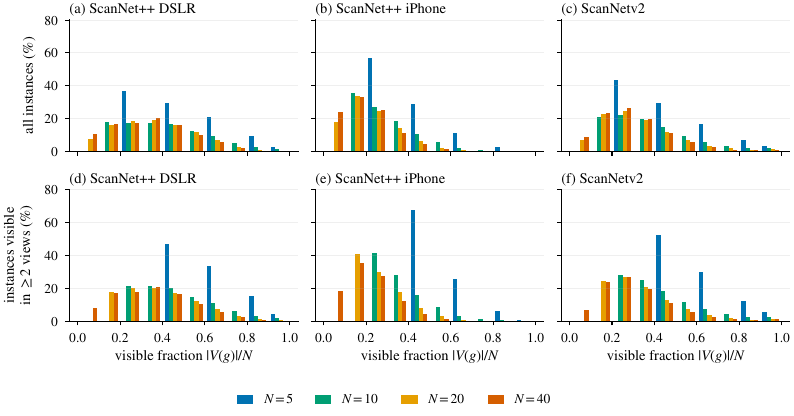}
\caption{Distribution of the fraction $\lvert V(g) \rvert / N$ of the input set in which a ground-truth instance is visible, at each $N$, over thing classes of the three captures. Top: all scored instances, bottom: instances visible in at least two views.}
\label{fig:coverage_dist}
\end{figure}

The dependence of $\varepsilon_{q}$ on $N$ in Table~\ref{tab:decomp} lies in the predictions rather than in the metric.
Whether a candidate is matched depends on the fraction of the input set in which the instance is visible and not on $N$ itself, and as $N$ grows more instances are visible in a small fraction of the set (Figs.~\ref{fig:coverage} to \ref{fig:coverage_dist}).

\begin{figure}[htp]
\centering
\includegraphics[width=\textwidth]{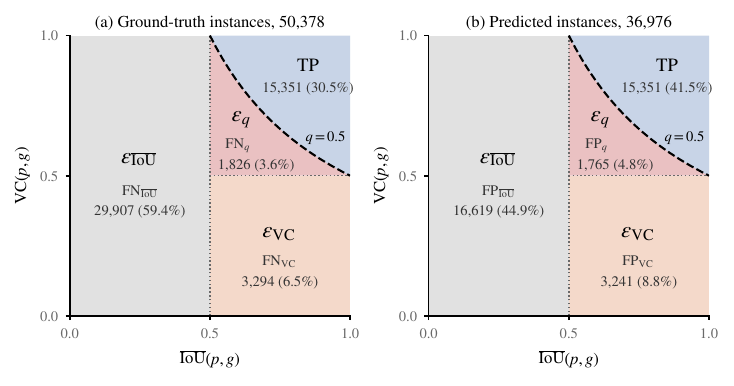}
\caption{The plane of the two factors of Eq.~\eqref{eq:q}, $\overline{\mathrm{IoU}}(p,g)$ and $\mathrm{VC}(p,g)$, divided by the three tests of Sec.~\ref{sec:vcpq_decomp}, for PanSt3R v2 on ScanNet++ at $N{=}10$ (thing classes): (a)~ground-truth instances, (b)~predicted instances. Each region is labelled with the term of Eq.~\eqref{eq:decomp} that its unmatched instances enter, with their number and share of all instances; the dashed curve is $q(p,g) = 0.5$. Matched pairs also enter $\varepsilon_{\overline{\mathrm{IoU}}}$ with $1 - \overline{\mathrm{IoU}}(p,g)$ and $\varepsilon_{\mathrm{VC}}$ with $\overline{\mathrm{IoU}}(p,g)\,(1 - \mathrm{VC}(p,g))$, which the regions do not show.}
\label{fig:decomp_regions}
\end{figure}

Fig.~\ref{fig:decomp_regions} shows the three tests as regions of the plane of $\overline{\mathrm{IoU}}(p,g)$ and $\mathrm{VC}(p,g)$ and places the unmatched instances of PanSt3R v2 behind Table~\ref{tab:decomp} at $N = 10$ in them.

Since $q(p,g)$ is the product of the two factors, the pairs with $q(p,g) = 0.5$ lie on the curve $\mathrm{VC}(p,g) = 0.5 / \overline{\mathrm{IoU}}(p,g)$, and the region between this curve and the two lines at 0.5 holds the pairs that pass both tests but not their product.
In the classification of unmatched instances (Sec.~\ref{sec:vcpq_decomp}), the test on $\overline{\mathrm{IoU}}(p,g)$ precedes the test on $\mathrm{VC}(p,g)$ because $\mathrm{VC}(p,g)$ compares supports alone and equals one for any two instances that are visible in the same views, so that a candidate whose mask does not overlap with $g$ carries no evidence about $g$, however consistent its support.
For instance, an instance with one candidate at $\overline{\mathrm{IoU}} = 0.2$, $\mathrm{VC} = 1$ and another at $\overline{\mathrm{IoU}} = 1$, $\mathrm{VC} = 0.5$ falls in $\mathrm{FN}_{\mathrm{VC}}$, whereas testing view consistency first would place it in $\mathrm{FN}_{\overline{\mathrm{IoU}}}$.
The candidates of an unmatched ground-truth instance are all predicted instances of the same class in the input set, matched or not, and likewise for an unmatched predicted instance with the roles of $p$ and $g$ exchanged.

\begin{figure}[htp]
\centering
\includegraphics[width=\textwidth]{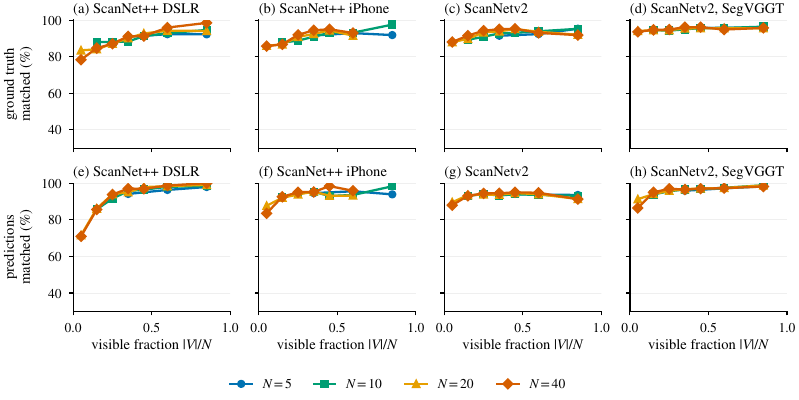}
\caption{As Fig.~\ref{fig:coverage}, for $\gamma = 1$.}
\label{fig:coverage_g1}
\end{figure}

Table~\ref{tab:decomp_all} gives the decomposition of Table~\ref{tab:decomp} over all classes, and its VC-PQ column is that of Table~\ref{tab:scores}.
SegVGGT predicts thing classes only and is given over thing classes alone, in Table~\ref{tab:decomp}.
Table~\ref{tab:decomp_counts} gives the counts behind Table~\ref{tab:decomp}, namely the ground-truth and predicted instances and the unmatched ones grouped by the first test that fails.
Table~\ref{tab:fpvc_support} counts the predicted instances of PanSt3R v2 that fail the view-consistency test by the number of views in which they are visible, where $\lvert V(p) \rvert$ counts the views in which the predicted mask, without its pixels on ground-truth void, has at least $\eta$ pixels.
Fig.~\ref{fig:coverage} and Fig.~\ref{fig:coverage_g1} give, for the instances that are visible in at least two views and have a same-class counterpart with $\overline{\mathrm{IoU}}(p,g) > 0.5$, the share that is matched as a function of the fraction $\lvert V \rvert / N$ of the input set in which the instance is visible, for $\gamma = 0$ and $\gamma = 1$.
Fig.~\ref{fig:coverage_dist} gives the distribution of that fraction over the ground-truth instances at each $N$.

\begin{table}[htp]
\centering\footnotesize
\caption{The decomposition of Table~\ref{tab:decomp} over all classes, for input sets of $N$ views (ScanNet++ pools the DSLR and iPhone runs). VC-PQ and the three terms sum to 100 in each block, and VC-PQ equals Table~\ref{tab:scores}. SegVGGT predicts thing classes only and is reported over thing classes alone (Table~\ref{tab:decomp}); EPS3D on ScanNetv2 is added here. --: as in Table~\ref{tab:scores}.}
\label{tab:decomp_all}
\setlength{\tabcolsep}{1.9pt}
\begin{tabular}{l cccc cccc cccc cccc}
\toprule
& \multicolumn{4}{c}{$N{=}5$} & \multicolumn{4}{c}{$N{=}10$} & \multicolumn{4}{c}{$N{=}20$} & \multicolumn{4}{c}{$N{=}40$} \\
\cmidrule(lr){2-5}\cmidrule(lr){6-9}\cmidrule(lr){10-13}\cmidrule(lr){14-17}
Method & VC-PQ & $\varepsilon_{\overline{\mathrm{IoU}}}$ & $\varepsilon_{\mathrm{VC}}$ & $\varepsilon_{q}$ & VC-PQ & $\varepsilon_{\overline{\mathrm{IoU}}}$ & $\varepsilon_{\mathrm{VC}}$ & $\varepsilon_{q}$ & VC-PQ & $\varepsilon_{\overline{\mathrm{IoU}}}$ & $\varepsilon_{\mathrm{VC}}$ & $\varepsilon_{q}$ & VC-PQ & $\varepsilon_{\overline{\mathrm{IoU}}}$ & $\varepsilon_{\mathrm{VC}}$ & $\varepsilon_{q}$ \\
\midrule
\multicolumn{17}{l}{\textit{ScanNet++}} \\
PanSt3R v1 & 22.0 & 72.8 & 4.0 & 1.1 & 19.2 & 73.6 & 4.6 & 2.6 & 17.9 & 73.7 & 4.1 & 4.2 & 17.4 & 73.6 & 3.7 & 5.3 \\
PanSt3R v2 & 24.5 & 68.8 & 5.1 & 1.5 & 22.2 & 67.6 & 7.0 & 3.1 & 20.7 & 67.0 & 7.1 & 5.2 & 20.0 & 66.6 & 6.6 & 6.8 \\
EPS3D & 2.4 & 96.8 & 0.6 & 0.2 & 2.0 & 97.1 & 0.5 & 0.3 & -- & -- & -- & -- & -- & -- & -- & -- \\
IGGT & 5.7 & 92.2 & 1.8 & 0.4 & 4.2 & 93.2 & 1.8 & 0.7 & -- & -- & -- & -- & -- & -- & -- & -- \\
\midrule
\multicolumn{17}{l}{\textit{ScanNetv2}} \\
PanSt3R v2 & 43.7 & 46.7 & 7.5 & 2.0 & 41.5 & 46.9 & 7.9 & 3.7 & 40.6 & 47.2 & 6.5 & 5.7 & 40.6 & 47.2 & 5.5 & 6.7 \\
EPS3D & 9.2 & 87.9 & 2.1 & 0.7 & 8.2 & 89.0 & 1.8 & 1.1 & -- & -- & -- & -- & -- & -- & -- & -- \\
\bottomrule
\end{tabular}
\end{table}

\begin{table}[htp]
\centering\footnotesize
\caption{Counts behind Table~\ref{tab:decomp} (thing classes): ground-truth and predicted instances, and the unmatched ones grouped by the first of the three tests, $\overline{\mathrm{IoU}}(p,g) > 0.5$, $\mathrm{VC}(p,g) > 0.5$, and $q(p,g) > 0.5$, that fails (Sec.~\ref{sec:vcpq_decomp}).}
\label{tab:decomp_counts}
\setlength{\tabcolsep}{4pt}
\begin{tabular}{l r rr rrr rrr}
\toprule
& & \multicolumn{2}{c}{instances} & \multicolumn{3}{c}{unmatched ground truth} & \multicolumn{3}{c}{unmatched predictions} \\
\cmidrule(lr){3-4}\cmidrule(lr){5-7}\cmidrule(lr){8-10}
Method & $N$ & GT & pred. & $\lvert\mathrm{FN}_{\overline{\mathrm{IoU}}}\rvert$ & $\lvert\mathrm{FN}_{\mathrm{VC}}\rvert$ & $\lvert\mathrm{FN}_{q}\rvert$ & $\lvert\mathrm{FP}_{\overline{\mathrm{IoU}}}\rvert$ & $\lvert\mathrm{FP}_{\mathrm{VC}}\rvert$ & $\lvert\mathrm{FP}_{q}\rvert$ \\
\midrule
\multicolumn{10}{l}{\textit{ScanNet++}} \\
PanSt3R v1 & 5 & 80{,}635 & 38{,}468 & 56{,}078 & 2{,}970 & 915 & 14{,}310 & 2{,}632 & 854 \\
 & 10 & 50{,}378 & 22{,}949 & 36{,}256 & 1{,}828 & 1{,}250 & 9{,}114 & 1{,}619 & 1{,}172 \\
 & 20 & 28{,}608 & 12{,}633 & 21{,}033 & 781 & 1{,}017 & 5{,}218 & 690 & 948 \\
 & 40 & 15{,}058 & 6{,}493 & 11{,}235 & 312 & 659 & 2{,}772 & 268 & 601 \\
PanSt3R v2 & 5 & 80{,}635 & 58{,}103 & 47{,}842 & 4{,}563 & 1{,}331 & 25{,}503 & 4{,}406 & 1{,}295 \\
 & 10 & 50{,}378 & 36{,}976 & 29{,}907 & 3{,}294 & 1{,}826 & 16{,}619 & 3{,}241 & 1{,}765 \\
 & 20 & 28{,}608 & 21{,}531 & 16{,}973 & 1{,}787 & 1{,}609 & 9{,}979 & 1{,}763 & 1{,}550 \\
 & 40 & 15{,}058 & 11{,}587 & 8{,}972 & 820 & 1{,}048 & 5{,}549 & 811 & 1{,}009 \\
EPS3D & 5 & 80{,}635 & 65{,}001 & 78{,}475 & 992 & 200 & 62{,}989 & 859 & 185 \\
 & 10 & 50{,}378 & 38{,}927 & 49{,}481 & 406 & 145 & 38{,}089 & 354 & 138 \\
IGGT & 5 & 80{,}635 & 40{,}029 & 71{,}204 & 2{,}315 & 649 & 30{,}764 & 2{,}199 & 599 \\
 & 10 & 50{,}378 & 21{,}514 & 45{,}937 & 1{,}113 & 572 & 17{,}158 & 1{,}063 & 537 \\
\midrule
\multicolumn{10}{l}{\textit{ScanNetv2}} \\
PanSt3R v2 & 5 & 73{,}066 & 72{,}143 & 28{,}767 & 6{,}477 & 1{,}669 & 29{,}117 & 5{,}294 & 1{,}579 \\
 & 10 & 44{,}726 & 46{,}158 & 17{,}365 & 3{,}798 & 2{,}065 & 19{,}481 & 3{,}250 & 1{,}929 \\
 & 20 & 24{,}831 & 26{,}717 & 9{,}508 & 1{,}470 & 1{,}810 & 11{,}618 & 1{,}341 & 1{,}715 \\
 & 40 & 12{,}721 & 13{,}915 & 4{,}867 & 473 & 1{,}111 & 6{,}140 & 439 & 1{,}066 \\
SegVGGT & 5 & 73{,}066 & 91{,}518 & 22{,}011 & 4{,}326 & 1{,}131 & 40{,}363 & 4{,}467 & 1{,}090 \\
 & 10 & 44{,}726 & 58{,}006 & 13{,}097 & 2{,}710 & 1{,}572 & 26{,}322 & 2{,}864 & 1{,}473 \\
 & 20 & 24{,}831 & 33{,}774 & 6{,}972 & 1{,}177 & 1{,}440 & 15{,}892 & 1{,}276 & 1{,}364 \\
 & 40 & 12{,}721 & 18{,}205 & 3{,}451 & 463 & 940 & 8{,}952 & 521 & 865 \\
\bottomrule
\end{tabular}
\end{table}

\begin{table}[htp]
\centering\small
\caption{Predicted instances that fail the view-consistency test ($\mathrm{FP}_{\mathrm{VC}}$) for PanSt3R v2 on ScanNet++ (thing classes), by the number of views $\lvert V(p)\rvert$ in which they are visible.}
\label{tab:fpvc_support}
\begin{tabular}{c r ccc}
\toprule
& & \multicolumn{3}{c}{share by $\lvert V(p)\rvert$ (\%)} \\
\cmidrule(lr){3-5}
$N$ & $\lvert\mathrm{FP}_{\mathrm{VC}}\rvert$ & 1 & 2 & $\geq 3$ \\
\midrule
5 & 4{,}406 & 77.4 & 19.2 & 3.4 \\
10 & 3{,}241 & 65.2 & 24.0 & 10.8 \\
20 & 1{,}763 & 45.1 & 28.1 & 26.8 \\
40 & 811 & 26.6 & 19.5 & 53.9 \\
\bottomrule
\end{tabular}
\end{table}

\FloatBarrier
\section{Factorization of VC-PQ}
\label{app:sqrq}

Table~\ref{tab:sqrq} gives VC-PQ and its two factors of Appendix~\ref{app:definitions} for every input set size, aggregated as in Table~\ref{tab:scores}.

\begin{table}[htp]
\centering\small\setlength{\tabcolsep}{3.4pt}
\caption{VC-PQ and its two factors VC-SQ and VC-RQ for input sets of $N$ views, all classes, aggregated as in Table~\ref{tab:scores}. The product $\text{VC-PQ} = \text{VC-SQ} \times \text{VC-RQ}$ holds for each class and not for the class averages shown here, and a class without a matched pair enters the average of VC-SQ with zero, as in PQ. --: as in Table~\ref{tab:scores}.}
\label{tab:sqrq}
\begin{tabular}{l l cccc cccc cccc}
\toprule
 & & \multicolumn{4}{c}{VC-PQ} & \multicolumn{4}{c}{VC-SQ} & \multicolumn{4}{c}{VC-RQ} \\
\cmidrule(lr){3-6}\cmidrule(lr){7-10}\cmidrule(lr){11-14}
Dataset & Method & $N{=}5$ & 10 & 20 & 40 & $N{=}5$ & 10 & 20 & 40 & $N{=}5$ & 10 & 20 & 40 \\
\midrule
ScanNet++ & PanSt3R v1 & 22.0 & 19.2 & 17.9 & 17.4 & 65.8 & 62.9 & 56.6 & 53.3 & 28.7 & 25.7 & 24.5 & 24.2 \\
 & PanSt3R v2 & 24.5 & 22.2 & 20.7 & 20.0 & 64.8 & 61.8 & 57.1 & 52.9 & 32.0 & 29.6 & 28.3 & 27.9 \\
 & EPS3D & 2.4 & 2.0 & -- & -- & 15.2 & 14.4 & -- & -- & 3.3 & 2.9 & -- & -- \\
 & IGGT & 5.7 & 4.2 & -- & -- & 36.4 & 32.4 & -- & -- & 7.4 & 5.7 & -- & -- \\
\midrule
ScanNetv2 & PanSt3R v2 & 43.7 & 41.5 & 40.6 & 40.6 & 78.9 & 76.6 & 74.7 & 73.7 & 54.9 & 53.5 & 53.6 & 54.2 \\
 & EPS3D & 9.2 & 8.2 & -- & -- & 47.2 & 45.8 & -- & -- & 12.2 & 11.2 & -- & -- \\
\bottomrule
\end{tabular}
\end{table}

\end{document}